\documentclass{article} % For LaTeX2e
\usepackage{iclr2026_conference,times}

\usepackage{amsmath,amsfonts,bm}

\def\eqref#1{equation~\ref{#1}}
\def\1{\bm{1}}

\DeclareMathAlphabet{\mathsfit}{\encodingdefault}{\sfdefault}{m}{sl}
\SetMathAlphabet{\mathsfit}{bold}{\encodingdefault}{\sfdefault}{bx}{n}

\definecolor{linkc}{rgb}{0, 0.44, 0.74}
\definecolor{eqc}{rgb}{1, 0, 0}
\definecolor{newcitecolor}{rgb}{0,0.6,0}
\usepackage[pagebackref=false,breaklinks=true,colorlinks=True,urlcolor=eqc,citecolor=linkc,linkcolor=eqc,bookmarks=false]{hyperref}

\usepackage[utf8]{inputenc} % allow utf-8 input
\usepackage[T1]{fontenc}    % use 8-bit T1 fonts
\usepackage{url}            % simple URL typesetting
\usepackage{booktabs}       % professional-quality tables
\usepackage{amsfonts}       % blackboard math symbols
\usepackage{nicefrac}       % compact symbols for 1/2, etc.
\usepackage{microtype}      % microtypography
\usepackage[dvipsnames,svgnames]{xcolor} 
\usepackage{subcaption}

\usepackage{marvosym}
\usepackage{graphicx}
\usepackage{caption}
\usepackage{multirow}
\usepackage{titletoc}
\usepackage{enumitem} 
\usepackage{colortbl}
\usepackage{booktabs}
\usepackage{algorithm}
\usepackage{float}

\usepackage{animate}
\usepackage{marvosym}
\usepackage{makecell}

\usepackage{color}

\usepackage{colortbl}
\usepackage{wrapfig}
\usepackage{csquotes}
\usepackage{amsmath}
\usepackage{algpseudocode}
\usepackage{hyperref}
\usepackage{url}
\definecolor{Red}{RGB}{192, 0, 0}
\definecolor{Blue}{RGB}{12, 114, 186} 
\usepackage{etoc}
\definecolor{myrefcolor}{rgb}{0, 0.367, 0.7}
\definecolor{Red}{RGB}{192, 0, 0}
\definecolor{Blue}{RGB}{12, 114, 186}

\newcommand{\ours}{{Follow-Your-Motion}}

\title{Follow-Your-Motion: Video Motion Transfer via Efficient Spatial-Temporal Decoupled Finetuning}

\author{
Yue Ma \textsuperscript{1}\textsuperscript{\dag},
Yulong Liu \textsuperscript{1}\textsuperscript{\dag},
Qiyuan Zhu \textsuperscript{1}\textsuperscript{\dag},
Xiangpeng Yang \textsuperscript{\S},
\textbf{Kunyu Feng},
Xinhua Zhang \textsuperscript{3},\\
\textbf{Zexuan Yan} \textsuperscript{5},
\textbf{Zhifeng Li} \textsuperscript{4},
\textbf{Sirui Han} \textsuperscript{1}\textsuperscript{\Letter},
\textbf{Chenyang Qi} \textsuperscript{1}\textsuperscript{\Letter},
\textbf{Qifeng Chen} \textsuperscript{1} \\[1mm]
\textsuperscript{1} HKUST,
\textsuperscript{2} HKUST(GZ),
\textsuperscript{3} Tsinghua University,
\textsuperscript{4} XIntelligence Technology Co., Limited
\textsuperscript{5} SJTU
}

\iclrfinalcopy % Uncomment for camera-ready version, but NOT for submission.
\begin{document}

\maketitle

\begingroup
\renewcommand{\thefootnote}{}
\footnotetext{\dag Equal contribution}
\footnotetext{\Letter\ Corresponding author}
\footnotetext{\S\ Project leader}
\endgroup

\begin{figure}[h]
%加一些变化大的
  \vspace{-1.2cm}

  \centering
  \includegraphics[width=0.95 \textwidth]{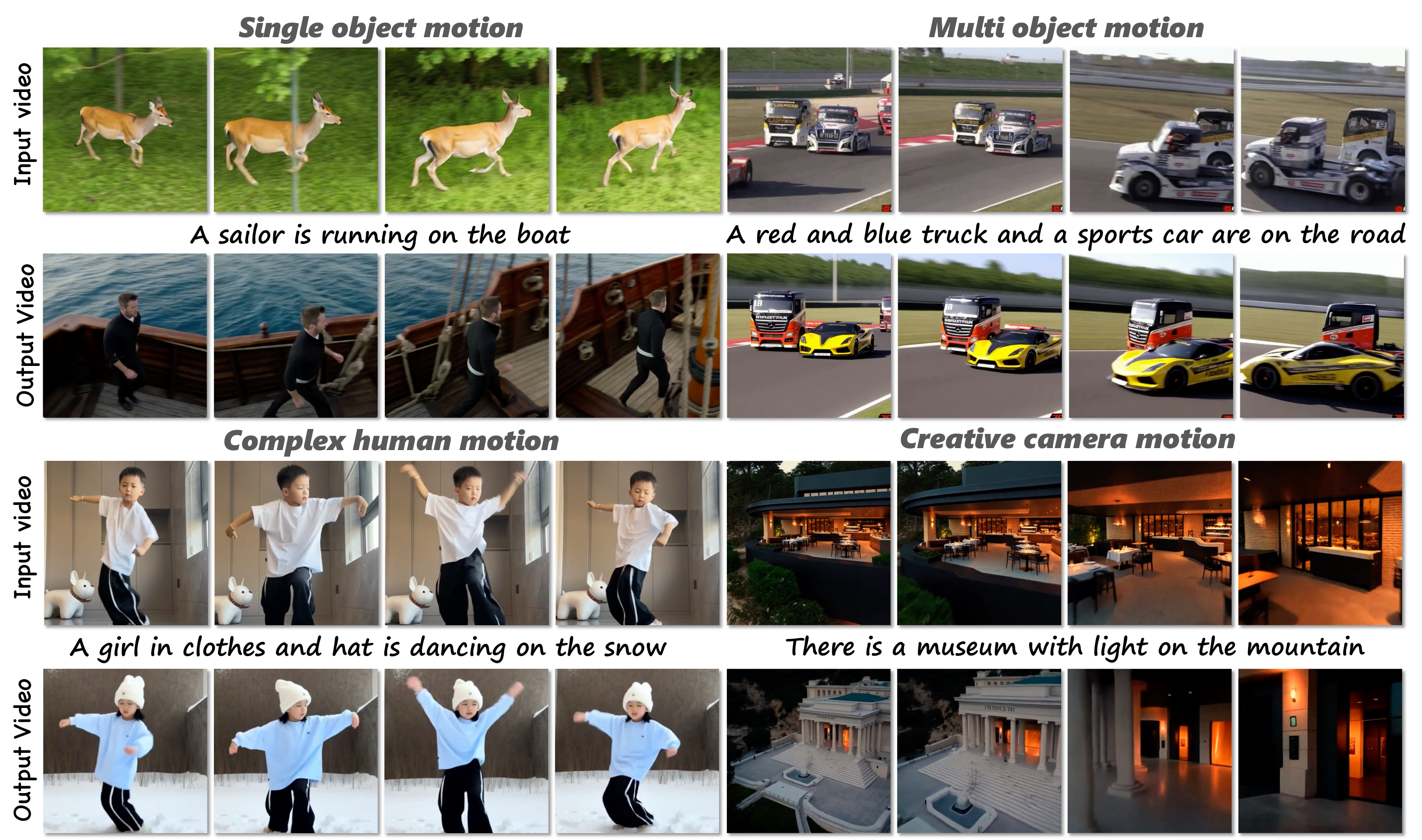} 

  \caption{\textbf{Showcases of our \ours}. Given an input video, \ours~enables generating the video with the same motion, including motion of single or multiple objects, complex poses of humans, and movements of the camera view.}

\label{teaser}
\end{figure}

\begin{abstract}
  % The abstract paragraph should be indented \nicefrac{1}{2}~inch (3~picas) on
  % both the left- and right-hand margins. Use 10~point type, with a vertical
  % spacing (leading) of 11~points.  The word \textbf{Abstract} must be centered,
  % bold, and in point size 12. Two line spaces precede the abstract. The abstract
  % must be limited to one paragraph.

  % Recently, breakthroughs in video modeling have allowed for controllable camera trajectories in generated videos.
  Recently, breakthroughs in the video diffusion transformer have shown remarkable capabilities in diverse motion generations. As for the motion-transfer task, current methods mainly use two-stage Low-Rank Adaptations (LoRAs) finetuning to obtain better performance. However, existing adaptation-based motion transfer still suffers from \textit{\textbf{motion inconsistency}} and \textit{\textbf{tuning inefficiency}} when applied to large video diffusion transformers. 
  Naive two-stage LoRA tuning struggles to maintain motion consistency between generated and input videos due to the inherent spatial-temporal coupling in the 3D attention operator. In addition, they require time-consuming fine-tuning processes in both stages. 
  To tackle these issues, we propose \ours, an efficient \textit{\textbf{three-stage}} video motion transfer framework that finetunes a powerful video diffusion transformer to synthesize complex motion.
  In \textit{\textbf{stage 1}}, we propose a spatial-temporal head classification technique to decouple the heads of 3D attention to distinct groups for spatial-appearance and temporal motion processing. We then finetune the spatial heads in the \textit{\textbf{stage 2}}. In the \textit{\textbf{stage 3}} of temporal head tuning, we design the sparse motion sampling and adaptive RoPE to accelerate the tuning speed. To address the lack of a benchmark for this field, we introduce MotionBench, a comprehensive benchmark comprising diverse motion, including creative camera motion, single object motion, multiple object motion, and complex human motion. We show extensive evaluations on MotionBench to verify the superiority of \ours.

  % learned motion is often coupled with the appearances in the training videos. 

  % the motion concept learned by these methods is often coupled with the limited appearances in the training videos
  % following the motion of input video, achieving the decouple of spatial and temporal 
  % We propose EffiVMT, a method for 

\end{abstract}

% \begin{abstract}
% The abstract paragraph should be indented 1/2~inch (3~picas) on both left and
% right-hand margins. Use 10~point type, with a vertical spacing of 11~points.
% The word \textsc{Abstract} must be centered, in small caps, and in point size 12. Two
% line spaces precede the abstract. The abstract must be limited to one
% paragraph.
% \end{abstract}

\section{Introduction}

Motion transfer aims to synthesize novel videos that faithfully replicate the motion dynamics,including camera movements and object trajectories from a given reference video. Unlike video-to-video translation methods~\citep{qi2023fatezero,wu2022tune}, which prioritize preserving low-level appearance and 2D spatial structure, motion transfer focuses exclusively on disentangling and reapplying motion patterns. This capability holds significant promise across diverse domains such as cinematic production, augmented reality, automated advertising, and social media content generation.

Recent advances in generative models have been dominated by diffusion models~\citep{rombach2022high}, which excel in producing high-fidelity visual content through stable optimization over Gaussian noise trajectories. The emergence of Diffusion Transformers (DiTs) has further elevated scalability in terms of model size, computational efficiency, and compatibility with large-scale video datasets. Leveraging pretrained video diffusion models, researchers have developed a spectrum of motion transfer techniques, broadly categorized into \textit{training-free} and \textit{tuning-based} paradigms.
% Training-free motion transfer methods~\citep{geyer2023tokenflow, pondaven2025ditflow,qi2023fatezero,xiao2024video, yang2025videograin} replace or optimize the generated motion representation with the reference video in inference time, while freezing the parameters of neural networks. \textit{e.g.},  SMM~\citep{yatim2024space} proposes a new descriptor by averaging video features along spatial dimensions.
% Motionshop~\citep{yesiltepe2024motionshop} uses the latent updates in the denoising pipeline as Motion Score representation for DiT models.
% % between the reference video and the output. 
% Although these training-free methods avoid the network training cost and generalize in both early UNet and recent DiT models, their generated motion is bottlenecked by the knowledge of the pretrained diffusion model.

Training-free approaches~\citep{geyer2023tokenflow, pondaven2025ditflow, chen2023attentive, qi2023fatezero, xiao2024video, yang2025videograin} operate entirely during inference by manipulating intermediate motion representations, such as attention maps or latent trajectories, without modifying model parameters. For instance, SMM~\citep{yatim2024space} introduces a spatially averaged feature descriptor to guide motion consistency, while MotionShop~\citep{yesiltepe2024motionshop} repurposes latent-space updates in the denoising process as a “Motion Score” for DiT models. Although these methods offer zero-training-cost generalization across both UNet and DiT architectures, their fidelity is inherently constrained by the motion priors embedded in the pretrained model.

\begin{figure*}[t]
    \centering
    \includegraphics[width=0.94\linewidth]{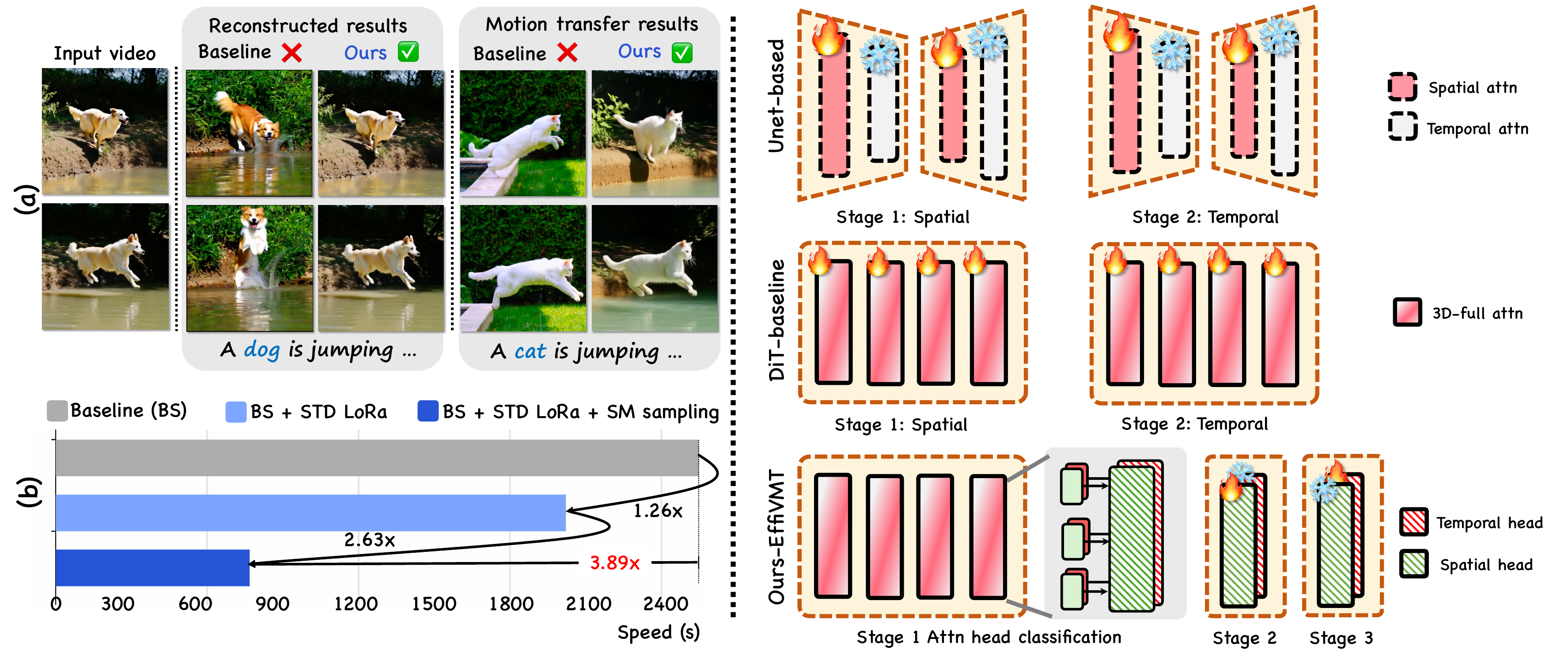}
    \caption{\textbf{Comparison between \ours~and baseline and Motivation}. \textbf{\textit{(a)\&(b):}} We finetune the baseline and our method 3,000 steps using Wan2.1~\citep{wan2025}. Our method gets better reconstruction and motion preservation. \textbf{\textit{(c):}} Despite the decoupling of temporal and spatial in UNet is common, applying it to modern DiT is still challenging because of its \textit{spatial-temporal mixed 3D full self-attention} blocks. To address it, we propose the spatial-temporal decoupled tuning for DiT, sparse motion sampling, and adaptive RoPE to synthesize video with complex motion efficiently. }
    \label{fig:compare_intro}
\end{figure*}

To overcome this limitation and capture complex, out-of-distribution motions, tuning-based methods~\citep{zhao2023motiondirector} optimize model parameters to explicitly encode reference motion.  In early UNet-based frameworks like MotionDirector (first row in the Figure~\ref{fig:compare_intro}(c).), temporal layers are fine-tuned independently to learn motion dynamics, while spatial layers remain frozen or jointly optimized. During inference, the learned motion is composited with the frozen model’s prior knowledge to generate novel videos. While effective, extending this paradigm to modern DiT architectures remains challenging due to their high computational cost and the entangled nature of spatial-temporal modeling in 3D self-attention blocks.

A naive baseline for DiT-based motion transfer involves applying Low-Rank Adaptation (LoRA) directly to all parameters within the 3D self-attention layers, as shown in the second row of Figure~\ref{fig:compare_intro}(c). More sophisticated methods, such as the approach proposed by \cite{abdal2025dynamic}, employ a two-stage spatial–temporal decoupled tuning strategy: first, spatial LoRAs are optimized on a subset of key frames to preserve appearance consistency; these are then frozen, and temporal LoRAs are tuned over the full video sequence to capture and transfer motion dynamics.
 % Tuning-based motion transfer, including U-Net-based \citep{zhao2023motiondirector} and DiT-based \citep{abdal2025dynamic} methods employ a spatial–temporal decoupled tuning paradigm: 
 % They first optimize a spatial LoRA on sampled frames to capture spatial structure, then freeze those weights and optimize a temporal LoRA over the full video frames to learn motion dynamics. 
 However, we argue that this two-stage procedure is inherently inefficient. Specifically, the limitations are listed as follows:

\begin{enumerate}[label=(\arabic*), nosep, leftmargin=*]
    \item \textit{\textbf{Motion inconsistency}}: 
     During the spatial tuning stage, both spatial and temporal attention heads are updated using static frames, inadvertently coupling spatial appearance with temporal dynamics. As shown in the top Fig.~\ref{fig:compare_intro}(a), for the naive baseline, both the reconstructed results and motion transfer results fail to follow the reference video. Therefore, tuning both spatial and temporal heads for appearance preservation is not reasonable.

    % In the first stage, both spatial and temporal heads in self-attention are tuned on a fixed frame, which leads to spatial-temporal coupling. As shown in the top Fig.~\ref{fig:compare_intro}, the naive baseline tunes both spatial and temporal heads, but both the reconstructed results and motion transfer results fail to follow the dog’s original motion. Therefore, tuning both spatial and temporal heads for appearance preservation is not reasonable.

    \item \textit{\textbf{Tuning inefficiency}}: 
     Recent analysis~\citep{xi2025sparse} reveals that 3D self-attention heads in DiTs naturally specialize, some focus on spatial relations, others on temporal coherence. Yet current methods indiscriminately tune all heads in each stage, resulting in parameter redundancy and suboptimal adaptation. Furthermore, since 3D VAEs inherently compress and interpolate temporal sequences, processing all reference frames during tuning ignores this latent interpolation capacity and introduces unnecessary computational overhead. 
    % Motivated by the observation in \citep{xi2025sparse}, the 3D full attention mechanism in DiT-based models contains attention heads that specialize in either spatial or temporal modeling. However, existing methods tune all attention heads in the spatial/temporal tuning stages, regardless of their roles, leading to inefficient parameter usage and redundancy. Moreover, since 3D VAEs compress temporal sequences into compact latent representations, using all reference frames for motion transfer ignores 3D VAEs' interpolation ability and introduces redundancy.
    
\end{enumerate}

% To address these challenges, we propose \ours, an efficient video motion transfer framework.
% Specifically, to mitigate the motion inconsistency problem, we first apply robust head matching to categorize attention heads into spatial and temporal types.
% During tuning, only spatial heads are updated in the spatial stage, and only temporal heads in the temporal stage.
% As shown in Figure~\ref{fig:compare_intro} (a), our method preserves the motion consistency in both reconstruction and transfer results compared to the naive baseline.
% To address tuning inefficiency, we introduce sparse motion sampling during the temporal tuning stage to accelerate. Furthermore, we propose adaptive RoPE to enhance the model’s ability to learn motion interpolation, allowing it to effectively capture motion dynamics even from sparse frames. As evidence in Figure~\ref{fig:compare_intro} (b), our decoupled tuning strategy reduces latency by $\textbf{1.26} \times$, and with sparse sampling, achieves a $\textbf{3.89}\times$ speed-up compared to tuning with all frames.

To tackle these challenges, we propose \ours, an efficient video motion transfer framework. First, to resolve motion inconsistency, we employ robust head matching to classify attention heads into spatial and temporal types. During tuning, spatial heads are updated only in the spatial stage, and temporal heads only in the temporal stage for preserving motion consistency in both reconstruction and transfer, as shown in Figure~\ref{fig:compare_intro}(a).   To improve tuning efficiency, we introduce sparse motion sampling during temporal tuning, significantly accelerating training. We further propose adaptive RoPE to enhance motion interpolation learning, enabling accurate motion capture even from sparse frames. As demonstrated in Figure~\ref{fig:compare_intro}(b), our decoupled strategy reduces latency by 1.26×, and with sparse sampling, achieves a 3.89× speed-up over full-frame tuning.

Together, these designs enable \ours ~ to generate high-fidelity videos that faithfully follow reference motion (See Figure~\ref{teaser}).  Additionally, to address the lack of benchmark in video motion transfer, we introduce MotionBench, which is a comprehensive benchmark covering single-object, complex human, multi-object, and camera motions across diverse scenes and styles. Our method outperforms existing baselines across various evaluation metrics, demonstrating its effectiveness in leveraging powerful DiTs for accurate motion transfer. Overall, our key contributions are summarized as follows:
\begin{itemize}
  \item We propose \ours, a three-stage motion transfer framework that efficiently adapts powerful video Diffusion Transformers (DiTs) to synthesize videos with complex, high-fidelity motion.

\item We identify and address two core challenges in DiT-based motion transfer: motion inconsistency and tuning inefficiency. To preserve motion coherence, we decouple spatial and temporal adaptation via specialized LoRA heads. To accelerate training, we introduce sparse motion sampling and adaptive RoPE for efficient yet accurate motion interpolation.

    \item To validate the effectiveness of our methods, we construct a benchmark MotionBench. We perform extensive experiments and user studies to evaluate our approach, which
     shows our method achieves state-of-the-art performance. 
    
\end{itemize}

\section{Related Work}

\noindent\textbf{Text-to-video generation.} 
% \subsection{Text-to-video generation} 
% Text-to-Video generation is a basic and popular research topic in computer vision. 
Text-to-video generation aims to produce realistic videos that precisely match the spatial visuals and temporal dynamics described in the input prompt.
To generate the complicated motion in the videos, 
% early methods are designed to handle the temporal dimension using Generative Adversarial Networks (GANs)~\citep{Treat2018GENERATIVEAN} or Variational Autoencoder(VAE)~\citep{}.  However, due to the challenges in training these models, they show the limited performance in general domain, \textit{e.g.,} animals.
% Following the success of diffusion model~\citep{Ho2020DenoisingDP} in image domain,  
diffusion-based video generation models~\citep{guoanimatediff, 
liu2025avatarartist, wang2025cinemaster, wang2026care, ma2026fastvmt, liu2025avatarartist, zhang2025magiccolor, liu2025longvideoagent, yang2025unified, shen2025follow, ma2026follow, ma2024followyourclick, chen2025contextflow, chen2024follow, long2025follow, ma2025controllable, lv2025bm, yang2025evctrl, chen2025s, chen2025taming, wang2025characterfactory, wang2025multishotmaster} are proposed to synthesize consistent results using a pretrained image diffusion model.
% such as LVDM~\citep{He2022LatentVD} and Modelscope~\citep{wang2023modelscope} 
Previous works~\citep{guo2023animatediff, chen2024m, wang2023modelscope} design the temporal module of UNet to generate consistent results.
% enhance the consistency of generated video clips.  
% Animatediff~\citep{} introduces the motion module and improves the performance by the
% pubic personalized text-to-video weight~\citep{guoanimatediff}. 
% Benefiting from the stability of diffusion-based model, these methods are scaled by a large dataset and show the impressive results on text-to-video generation. 
Recently, the emergence of Diffusion Transformer-based methods for text-to-video generation has exhibited superior performance in quality and consistency. These powerful scaling transformers, including Sora~\citep{Liu2024SoraAR}, CogVideoX~\citep{Yang2024CogVideoXTD}, EasyAnimate~\citep{xu2024easyanimate}, HunyuanVideo~\citep{kong2024hunyuanvideo}, and Wan2.1~\citep{wan2025}, enable generating more realistic video clips from given detailed prompts, paving the way for various downstream video generation tasks. 
\noindent\textbf{Video Motion transfer.} 
Motion transfer involves an important demand: creating a novel video and maintaining the motion from the reference one.  
% Different with video-to-video translation~\citep{zhao2023controlvideo},  the target of motion transfer is to decouple the spatial appearance and temporal motion from input video.
Some methods leverage the explicit control signal~\citep{ma2024follow, xue2024follow, feng2025dit4edit, wang2024cove, zhu2025multibooth, zhu2024instantswap, xing2024make, ma2023magicstick,  xing2024dynamicrafter, zhang2025framepainter,ma2024followyouremoji, zhao2023controlvideo}
% , such as poses~\citep{}, depths~\citep{}, bounding boxes~\citep{}, trajectories~\citep{} and motion masks~\citep{}, 
to achieve motion transfer from the reference video. However, these methods rely on a huge control signal dataset and cost large computational resources. Thanks for the powerful pretrained text-to-video generation model, the researchers pay attention to motion transfer using implicit control, including training-free or tuning-based paradigm.
% Zero-shot motion transfer methods, such as TokenFlow~\citep{}, Rerender-a-Video~\citep{} and RAVE~\citep{}, mainly leverage inversion strategy to achieve precise control while maintaining visual quality. 
% For training-free methods, like DiTFlow~\citep{pondaven2025ditflow}, MotionShop~\citep{yesiltepe2024motionshop}, and MotionMaster~\citep{hu2024motionmaster}
For training-free methods~\citep{hu2024motionmaster, pondaven2025ditflow, yesiltepe2024motionshop},  they extract a motion embedding in the inference stage and use the gradient to guide optimization.  However, these methods fail to transfer the complex motion. For tuning-based methods, they~\citep{jeong2024dreammotion, zhao2023motiondirector} always fine-tune model parameters to utilize different attention for temporal and spatial information. Current works~\citep{jeong2024dreammotion, zhao2023motiondirector} employ the dual-path LoRA structure to separate motion and appearance. 
% DreamVideo~\citep{jeong2024dreammotion} and Customize-A-Video~\citep{ren2024customize} further improve the separation stratage by distinct branches. 
However, these methods are developed on the UNet-based pretrained model~\citep{chai2023stablevideo}, making them unsuitable for DiTs.  In contrast, our proposed method is the first one-shot DiT-based motion transfer framework. Using the video diffusion transfer as the foundation model, our method extends the boundary of motion transfer performance.

\vspace{-1mm}

% MotionDirector (MD) [40] made a significant advancement with its innovative dual-path LoRA architecture, effectively separating motion and appearance characteristics through specialized components that enable precise control over temporal dynamics. DreamVideo [31] and Customize-A-Video [22] further refined this separation using distinct branches for appearance and motion learning.

% motion transfer approaches aim for complete disentanglement of the original video structure, focusing on motion alone. Some methods use training to condition on motion signals like trajectories, bounding boxes and motion masks [8, 11, 56, 57, 60, 63, 64], but this implies significant costs. Other approaches train motion embeddings [23] or finetune model parameters [15, 22, 58, 65]. However, these methods use separate attention for temporal and spatial information, making them unsuitable for DiTs. Optimization-based approaches extract a motion representation at inference [14, 21, 59, 62], which is more suitable for cross-architecture applications. TokenFlow [14] has a nearest-neighbor based approach on diffusion features, employing expensive sliding window analysis. SMM [62] employ spatial averaging, while MOFT [59] discover motion channels in diffusion features.

% \subsection{Low-rank adaptation finetuning.}

\section{Method}

\begin{wrapfigure}{r}{0.45\textwidth}
    \vspace{-1em} % 可选：调整顶部间距
    \begin{minipage}{\linewidth}
        \begin{algorithm}[H]
            \SetAlgoNoLine
            \DontPrintSemicolon
            \caption{Dual attention decoupling}
            \label{alg:head_classification}
            \KwIn{$Q,K \in \mathbb{R}^{H \times S \times D}$: query and key where $S = F \times H \times W$}
            \KwOut{Closest head type: $t_{\text{head}}$}

            \BlankLine
            \small{\text{$\triangleright$ Target spatial \& temporal attention maps:}}
            \small{\text{ $[\text{head}, S, S]$}}

            $M_{\text{spatial}} \leftarrow$ gen\_spatial\_maps($F, H, W$) \\
            $M_{\text{temporal}} \leftarrow$ gen\_temporal\_maps($F, H, W$)

            \BlankLine
            \small{\text{$\triangleright$ Get attention maps of input data: $[\text{head}, S, S]$}}

            $M_{\text{input}} \leftarrow$ Softmax($Q \cdot K^\top / \sqrt{D}$)

            \BlankLine
            \text{$\triangleright$ Calculate similarity metrics}

            $\text{Sim}_s \leftarrow \left\|M_{\text{input}} \odot M_{\text{spatial}}\right\|_{\text{mean}}$ \tcp*{mean over (1,2)}
            $\text{Sim}_t \leftarrow \left\|M_{\text{input}} \odot M_{\text{temporal}}\right\|_{\text{mean}}$

            \BlankLine
            \text{$\triangleright$ Classify head type: Boolean tensor $[\text{head}]$}

            $t_{\text{head}} \leftarrow (\text{Sim}_s < \alpha \cdot \text{Sim}_t)$
        \end{algorithm}
    \end{minipage}
    \vspace{-1em} % 可选：调整底部间距
\end{wrapfigure}

% As proposed in previous work~\citep{abdal2025dynamic,zhao2023motiondirector}, a naive baseline (Fig.~\ref{fig:compare_intro}) of motion transfer is to first optimize the LoRA weight for sampled video frames as a text-to-image task, where the LoRA obtained in this stage is called spatial LoRA ($\Delta W_s$). In the next stage, consecutive multiple frames are treated as ground truth to optimize the temporal LoRA ($\Delta W_t$), while the spatial LoRA is loaded and kept frozen. During the inference stage, only temporal lora ($\Delta W_t$) is loaded to transfer the video motion in the source video while presenting the appearance bias leakage. Although this strategy can generate results for the UNet-based diffusion model, the computation cost for the DiT-based video diffusion model is still \textit{too expensive}. As shown in Fig.~\ref {fig:compare_intro}, such naive LoRA tuning cannot reproduce the motion in the input video after tuning 3000 steps (3042 seconds on a single H20 GPU). 
Following prior work~\citep{abdal2025dynamic,zhao2023motiondirector}, a naive baseline first optimizes spatial LoRA weights ($\Delta W_s$) by treating sampled frames as independent text-to-image instances. 
Subsequently, temporal LoRA ($\Delta W_t$) is learned by fine-tuning on consecutive frame sequences while freezing $\Delta W_s$. 
At inference, only $\Delta W_t$ is applied to transfer motion. However, this leads to appearance leakage and remains computationally expensive for DiT-based video diffusion models. 
As shown in Fig.~\ref{fig:compare_intro}, even after 3,000 optimization steps (3,042s on a single H20 GPU), motion fidelity is unsatisfactory.

Previous naive LoRA tuning faces two main challenges. (1) Recent Video DiT models leverage 3D attention block without explicit temporal blocks, which makes it difficult to disentangle temporal parameters, and fine-tuning LoRA on whole attention parameters results in larger parameter number (\textit{e.g.}, 29.5 M for naive LoRA) (2) Finetuning on multiframe videos increases token sequence length (\textit{e.g.}, 24276 tokens for 81 frames) and computation cost. 
% In the following sections, we propose head analysis techniques to sparsify the 3D attention block and accelerated tuning strategies for temporal motion learning.

% To address these challenges, in Sec~\ref{sec: head_classification}, we first present an attention head classification strategy to decouple the spatial and temporal model parameters by analyzing the sparsity of attention in the pretrained video diffusion transformer. Then we introduce an efficient and economical tuning framework (Sec.~\ref{sec: decoupled training}) for learning spatial appearance and temporal motion in the source video. Note that we use WAN as our pretrained backbone in our experiments, but our method can generalize to other video DiT models.
To address these challenges, we first propose an attention head classification strategy (Sec.~\ref{sec: head_classification}) that decouples spatial and temporal parameters by analyzing attention sparsity in the pretrained Video DiT. 
Building on this, we introduce an efficient tuning framework (Sec.~\ref{sec: decoupled training}) to separately learn spatial appearance and temporal motion from the source video. 
While we use WAN as the pretrained backbone in our experiments, our method is model-agnostic and readily generalizes to other Video DiT architectures.

\begin{figure*}[tbp]
  \centering
  % \begin{subfigure}[t]{0.5\textwidth}
    \includegraphics[width=0.94\linewidth]{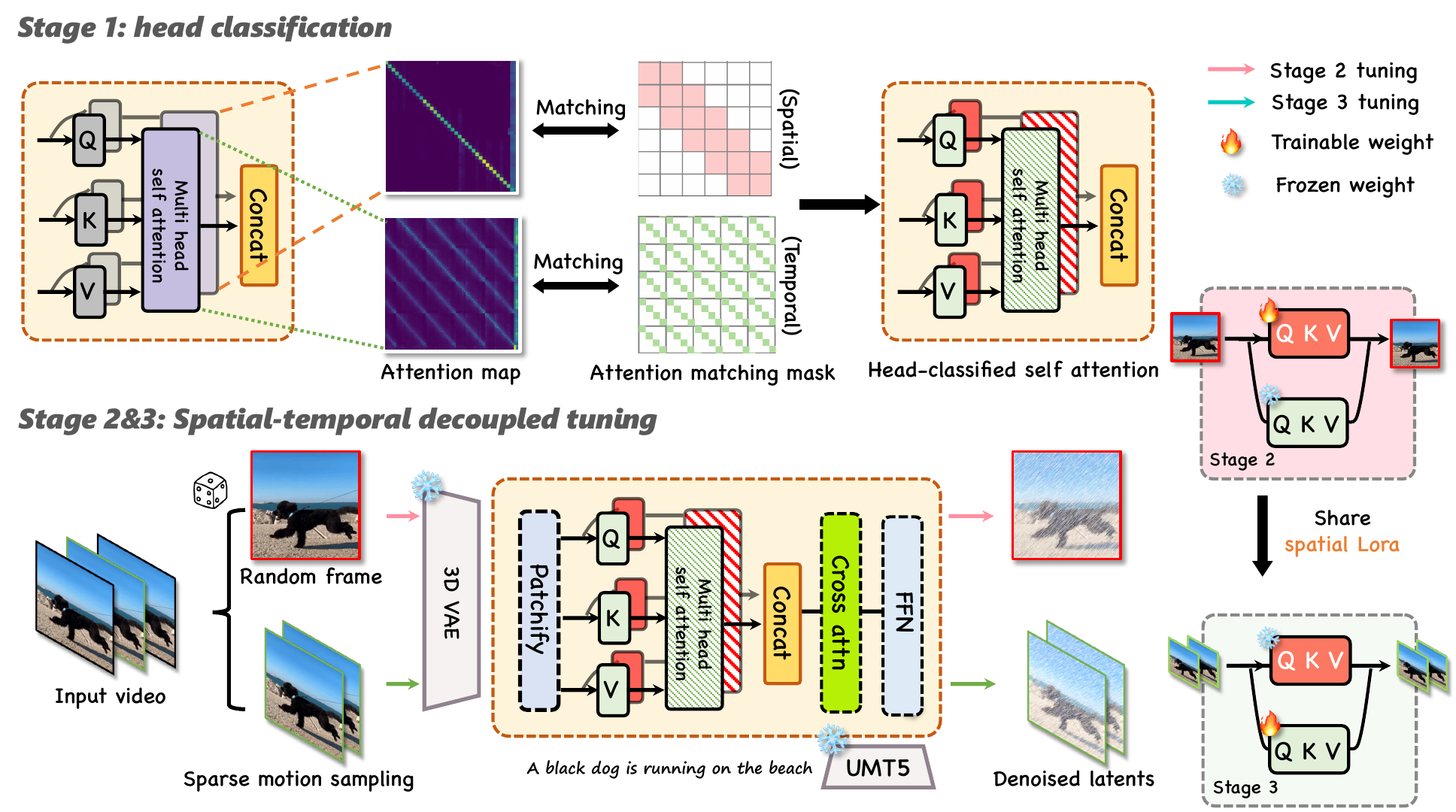}
    \caption{\textbf{Overview of our methods.} 
    \textit{\textbf{Stage 1}}: We first classify the attention heads using a pseudo spatial attention map. \textit{\textbf{Stage 2}}: After attention classification, we first tune the spatial LoRA using a random frame in the video. \textit{\textbf{Stage 3}}: After finishing spatial LoRA tuning, we load the spatial LoRA weight and conduct temporal tuning using sparse motion sampling and adaptive RoPE.}
  % \end{subfigure}
  \vspace{-1em}
\end{figure*}

\subsection{Stage 1: Spatial-temporal attention classification}
\label{sec: head_classification}
The pretrained video DiT model Wan utilizes unified 3D attention instead of separated spatial and temporal attention, which brings challenges to motion information decoupling~\citep{pondaven2025ditflow}, training efficiency, and storage cost~\citep{pondaven2025ditflow}. Inspired by evidence in previous work~\citep{xi2025sparse}, we leverage the inherent sparsity in 3D Full Attention of video DiT to decouple the parameters for temporal motion and spatial appearance.

\noindent\textbf{Dual attention decoupling.} 
% As shown in Alg.~\ref{alg: head classification}, our method first classifies the attention heads in Wan to either the temporal head or the spatial head. Givens input query and key tokens $Q, K \in \mathcal{R}^{H\times S\times D}$ with head number $H$, sequence length $S$ and dimension $D$, we create a pseudo spatial attention map ground truth $M_{spatial}\in \mathcal{R}^{H\times S \times S}$. If the query token and key token are at the same frame but different coordinates, we assign their corresponding value in attention maps as $1$, otherwise, the attention map value is $0$. In the same way, we create pseudo temporal attention map $M_{temp}$ where the query and token in the same coordinates but different frame number corresponds to $1$ in the attention map. Next, we calculate the attention map $M_{input}$ at each layer and each head for the input video, to get its cosine similarity $\text{Sim}_{s}$ for $M_{spatial}$, and $\text{Sim}_{t}$ for $M_{temporal}$. We set the head type as temporal head if $\text{Sim}_{s}< \alpha \text{Sim}_{t}$, where $\alpha$ is empirically set as $1.3$ to balance the spatial and temporal heads number.
As shown in Alg.~\ref{alg:head_classification}, our method classifies attention heads in Wan into temporal or spatial types. We take query and key tokens $Q, K \in \mathbb{R}^{H \times S \times D}$ as input, where $H$ is the number of heads, $S$ is the sequence length, and $D$ is the feature dimension. 

We prepare pseudo ground truths: for \textcolor{BrickRed}{spatial attention map}, $M_{\text{spatial}}[i, j] = 1$ if points \((i, j)\) are near the main diagonal ( within a predefined range), otherwise $0$; for \textcolor{teal}{temporal attention map}, \(M_{\text{temporal}}[i, j] = 1\) if points \((i, j)\) are near diagonals parallel to the main diagonal (identical spatial positions in different frames), otherwise $0$. 
% The number of $1s$ in \(M_{\text{spatial}}\) and \(M_{\text{temporal}}\) are equal.  

We compute the cosine similarity \(\text{Sim}_s\) between the input attention map \(M_{\text{input}}\) and \(M_{\text{spatial}}\), and \(\text{Sim}_t\) between \(M_{\text{input}}\) and \(M_{\text{temporal}}\). A head is classified as temporal if \(\text{Sim}_s < \alpha \cdot \text{Sim}_t\), where \(\alpha = 1.25\)(empirically set to balance the number of spatial and temporal heads).

\begin{figure*}[t]
  \centering
  % 左半：图片
  \begin{minipage}[tl]{0.51\textwidth}
    \includegraphics[width=\linewidth]{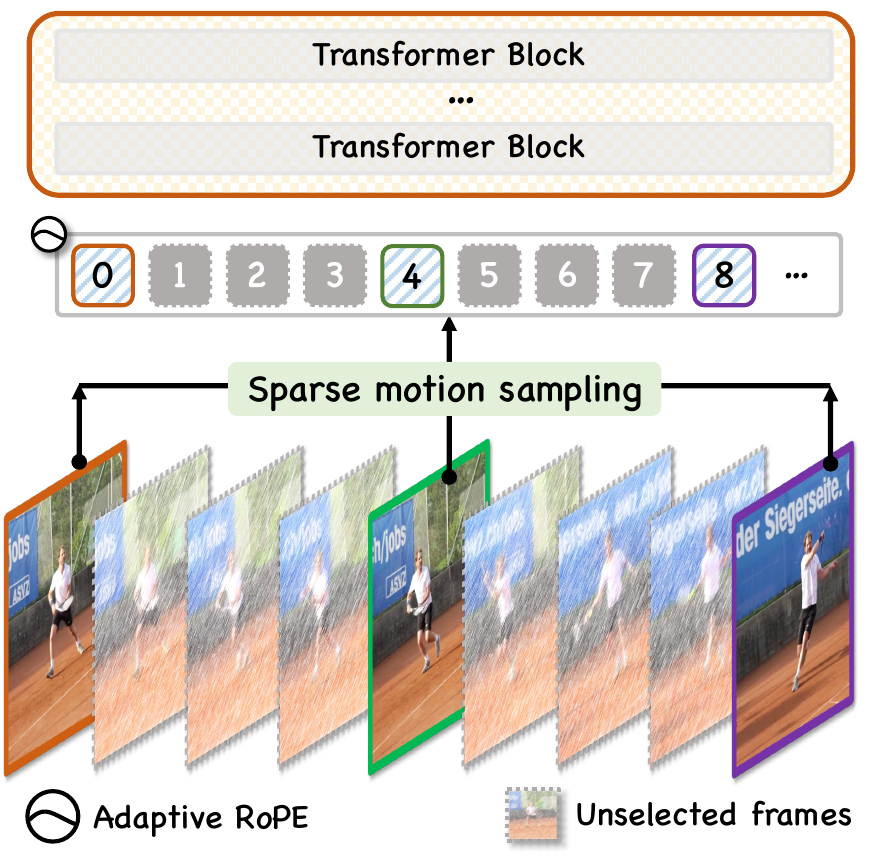}
    \caption{\textbf{Illustration of sparse motion sampling and adaptive RoPE}. The adaptive RoPE is utilized to represent frame position in the video.}
    \label{fig:abla_merge}
   \end{minipage}
  \hfill
  % 右半：算法
  % \vspace{-0.6cm}
  % （可选）统一总 caption：
  % \caption{（左）随机采样消融；（右）Dual Attention Fusion 算法。}
  % \label{fig:abla_and_alg}
% \end{figure} 
\begin{minipage}[tr]{0.45\textwidth}
\begin{algorithm}[H]
\hspace{3pt}
  \renewcommand{\thealgocf}{2}
  \SetAlgoNoLine
  \DontPrintSemicolon
  \caption{Dual Attention Fusion}\label{alg:dual_attention}
  \KwIn{
    \begin{tabular}{@{}l@{}}
      $x \in \mathbb{R}^{H \times S \times D}$: input sequence \\
      $f_{\mathrm{regs}}$: positional frequencies \\
      $d_{\mathrm{temp}}$: temporal dimension size
    \end{tabular}
  }
  \KwOut{Fused output $y \in \mathbb{R}^{H \times S \times D}$}
  \BlankLine
  \text{ $\triangleright$ Channel concatenate, and normalize}\\
  $Q \leftarrow \mathrm{Norm}([q_{\mathrm{temp}}(x)\|q_{\mathrm{spat}}(x)])$\\
  $K \leftarrow \mathrm{Norm}([k_{\mathrm{temp}}(x)\|k_{\mathrm{spat}}(x)])$\\
  $V \leftarrow [v_{\mathrm{temp}}(x)\|v_{\mathrm{spat}}(x)]$\\
  \BlankLine
  \text{ $\triangleright$ Rotary Position Embeddings for Q, K}\\
  $\widetilde{Q} \leftarrow \mathrm{RoPE}(Q, f_{\mathrm{regs}}, H)$\\
  $\widetilde{K} \leftarrow \mathrm{RoPE}(K, f_{\mathrm{regs}}, H)$\\
  \BlankLine
  \text{ $\triangleright$ Multi-Head Attention}\\
  $x \leftarrow \mathrm{Attention}(\widetilde{Q},\,\widetilde{K},\,V;\,H)$\\
  \BlankLine
  \text{ $\triangleright$ Dual Output Projection Fusion}\\
  % $y \leftarrow o_{\mathrm{temp}}(x[:,:,:\!d_{\mathrm{temp}}]) \;+\;
              % o_{\mathrm{spat}}(x[:,:,d_{\mathrm{temp}}:\!])$\\
  \small{$y \leftarrow o_{\mathrm{temp}}(x[:d_{\mathrm{temp}}])+
              o_{\mathrm{spat}}(x[d_{\mathrm{temp}}:])$}\\
  \Return{$y$}
\end{algorithm}
\end{minipage}
\end{figure*}

\noindent\textbf{Dual attention fusion.}
Then, we rearrange the channels of the linear layers in full 3D attention $q, k, v, o$ to two parallel branches for temporal attention and spatial attention. The forward algorithm of a single rearranged block is shown in Alg.~\ref{alg:dual_attention}. Given input sequence $x$, we concatenate the features from the temporal and spatial branches along the channel dimensions to get the tokens of query $Q$, key $K$, and value $V$. After applying rotary position embedding and scaled dot product attention, feature $x$ is split along the channel dimension, and fed to $\mathrm{o}_{temp}$ and $\mathrm{o}_{spat}$. Finally, the summed feature is returned at the end of the attention block.  

\subsection{Stages 2\&3: Spatial-temporal decoupled tuning}
\label{sec: decoupled training}
% learnable parameter, equation annotation should be same as stage 0
% Equation 3.1.4 image lora training algorithm
\noindent\textbf{Spatial LoRA tuning.}
As the parameters of the attention block are decoupled in the previous stage, we can use the spatial and temporal branches in two stages to learn the appearance and motion in the reference videos, respectively. Following previous work~\citep{abdal2025dynamic,zhao2023motiondirector}, we first inject LoRAs $\theta_{spat}$ into the spatial heads branch $(q_{spat}, k_{spat}, v_{spat}, o_{spat})$ to learn the spatial appearance in stage 2. In each iteration, We randomly sample a single frame $x_i$ from index $\{0, 1, 2, ..., F-1\}$, and optimize the spatial LoRA $\theta_{spat}$ as a text-to-image model using the training loss:
\begin{equation}
\label{eq:image_tune}
\mathcal{L}_{spat} = E_{x_{i,1}\sim P_{data}, x_{i,0} \sim N(0, I), i \sim U(0, F)}\left\|v_{i,t}-v_{\theta_{spat}}\left(x_{i,t}, t, p \right)\right\|_2^2,
\end{equation}
where $t$ is time step, $p$ is positional embedding and $v$ represents the velocity in the diffusion model. 

\noindent\textbf{Temporal LoRA tuning.}  Once the spatial LoRA $\theta_{spat}$ gets converged, we freeze $\theta_{spat}$ in the model, and continue to finetune temporal LoRA parameters $\theta_{temp}$ of temporal heads branch $(q_{temp}, k_{temp}, v_{temp}, o_{temp})$. Since the Wan~\citep{wan2025} is pretrained on a large frame number $F=81$, fine-tuning on the original number $F=81$ costs too expensive computation~(Fig.~\ref{fig:compare_intro}). To alleviate the high computation requirement for videos, we propose the \textit{\textbf{sparse motion sampling}}, which finetune our temporal LoRA ${\theta_{temp}}$ on a sampled video with fewer frame number $F_{samp}=17$ and then infer with the original frame number. While recent transformer models apply Rotary Positional Embedding (RoPE)~\citep{vaswani2017attention} to encode the relative position dependency according to the frame index, sampling frames from $F$ to $F_{samp}$ breaks the original dependency and thus deteriorates the motion quality. Motivated by previous text-to-image DiT models~\citep{kong2024hunyuanvideo, yang2024cogvideox}, we propose the \textit{\textbf{adaptive RoPE}}, a centralized scaling positional encoding along the frame index to align the position range with different total frame numbers. For each frame with temporal index $i \in [0, 1, ..., {F_{samp}-1}]$, its temporal positional embedding is assigned as:
\begin{equation}
\label{eq:temp_debias2}
\text{PE}_{x_{i}} = f(\frac{F}{2} + \frac{F}{F_{samp}}(i-\frac{F_{samp}}{2})),
\end{equation}

which ensures that videos with less frame number $F_{samp}$ have the same input range $[0, F]$ for the embedding function $f$, as the pertaining stage of video DiT.

\vspace{-2mm}

To further decouple the temporal motion from spatial appearance, we further introduce a motion loss~\citep{ling2024motionclone} by eliminating the appearance and focusing on the changes in the temporal dimensions. We first define the motion latent $\hat{v}$ for each frame $i$ as: $\hat{v}_{i,t} = v_{i,t} - v_{i-1,t}.$
% \end{equation}

Then, we define the motion loss following~\citep{zhao2023motiondirector} as the negative cosine similarity between the ground truth motion latent and predicted motion latent:
\begin{equation}
\label{eq:temp_debias}
\mathcal{L}_{Motion} = E_{x_{i,1}\sim P_{data}, x_{i,0} \sim N(0, I), i \sim U(0, F)}[1- \text{CosineSim} (\hat{v}_{i,t}, \hat{v}_{\theta_{temp}}\left(x_{i,t}, t, p \right))].
\end{equation}

Finally, the total loss for temporal LoRAs is the combination of general video denoising loss and motion loss as $\mathcal{L}_{temp}$ = $\mathcal{L}_\text{{video\_denoise}}$ + $\mathcal{L}_{Motion}$.

% Equation 3.1.5 attention debias loss 
% MotionDirector (video frame - random sample (image frame))
% Ours (video frame - neighbor frame)
% To ask qiyuan loss curve or visual comparison; choose video with large motion

% Video frame number sampling 80 -> 16 ; add rope rescaling (single row figure visualization)

% Visual Exp results should be at head of page 6 or 7
\begin{figure*}[t]
  \centering
  \includegraphics[width=0.94\linewidth]{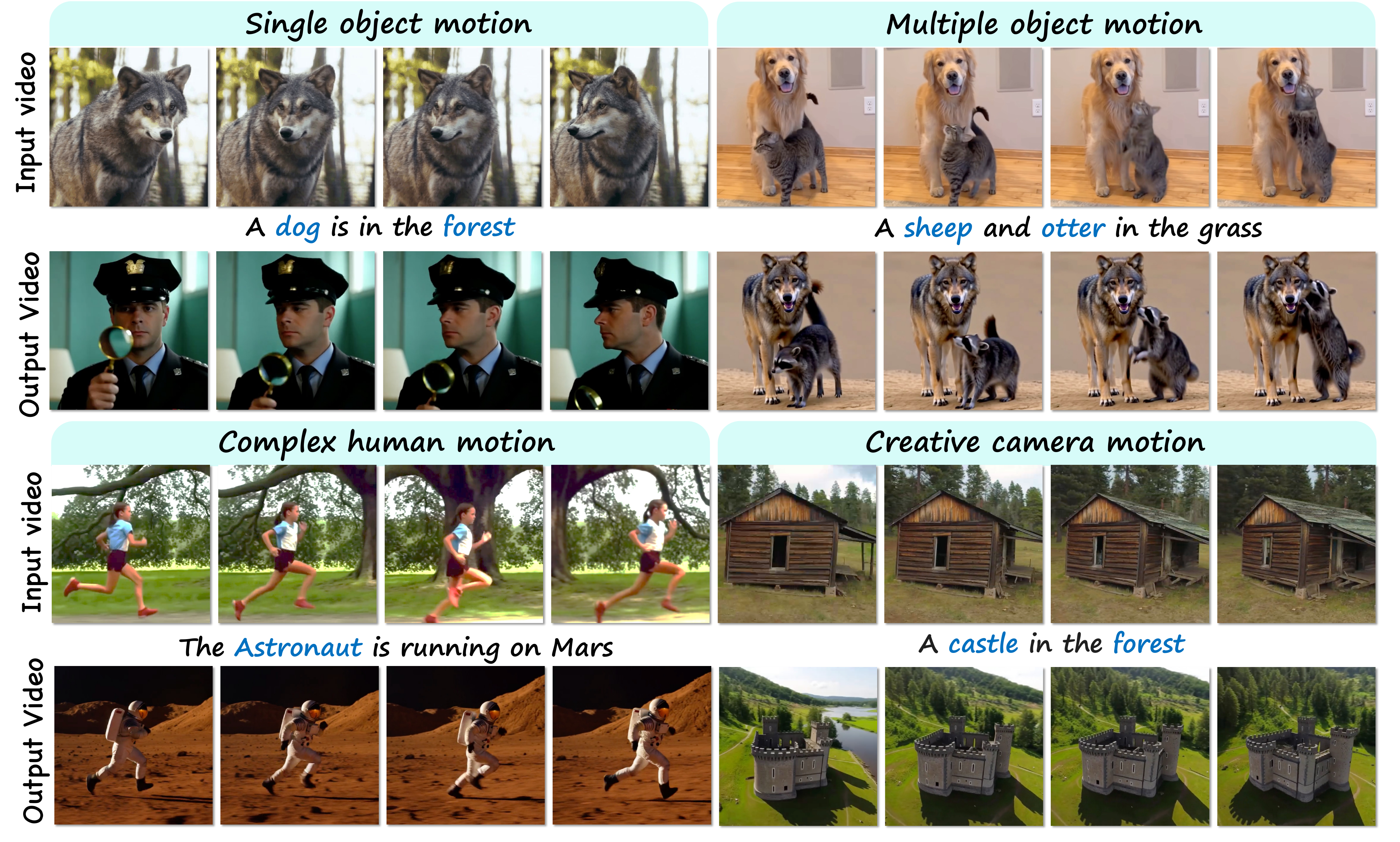}
  \caption{\textbf{Gallery of our proposed methods.} Given a reference video, our \ours~capability of generating a high-quality video clip with the same motion, including single object motion, multiple object motion, complex human motion, and camera motion. }
  \label{fig:gallery}
  % \vspace{-0.6cm}
\end{figure*}
% \vspace{-0.6cm}

\section{Experiments}
\label{sec: experiment}
% In this section, we first show more cases in Fig.~\ref{fig:gallery} and describe the implementation details in Sec.~\ref{sec:details}. Then we show the qualitative and quantitative results in Sec.~\ref{sec: comparison}. Finally, in Sec.~\ref{sec:ablation}, the ablation studies about the proposed modules are conducted to evaluate the effectiveness. We also introduce the details about MotionBench in the Appendix.

\subsection{Implementation details}
\label{sec:details}
In our experiment, we employ the open-sourced video generation model WAN-2.1~\citep{wan2025} as the base text-to-video generation model. The LoRA ranks are 128 in both stages. 
We first randomly select a single frame and take about 3,000 steps for spatial appearance learning. The AdamW~\citep{loshchilov2017decoupled} optimizer is utilized, and the learning rate is $1 \times 10^{-5}$. The spatial weight decay is $0.1$. 
During the third tuning stage, we freeze the spatial head LoRA and only train the temporal head LoRA for 2000 steps with learning rate $1 \times 10^{-5}$ and weight decay $0.99$.  
% More details and evaluation metrics can be found in Appendix.

\begin{figure*}[t]
  \centering
  \includegraphics[width=0.94\linewidth]{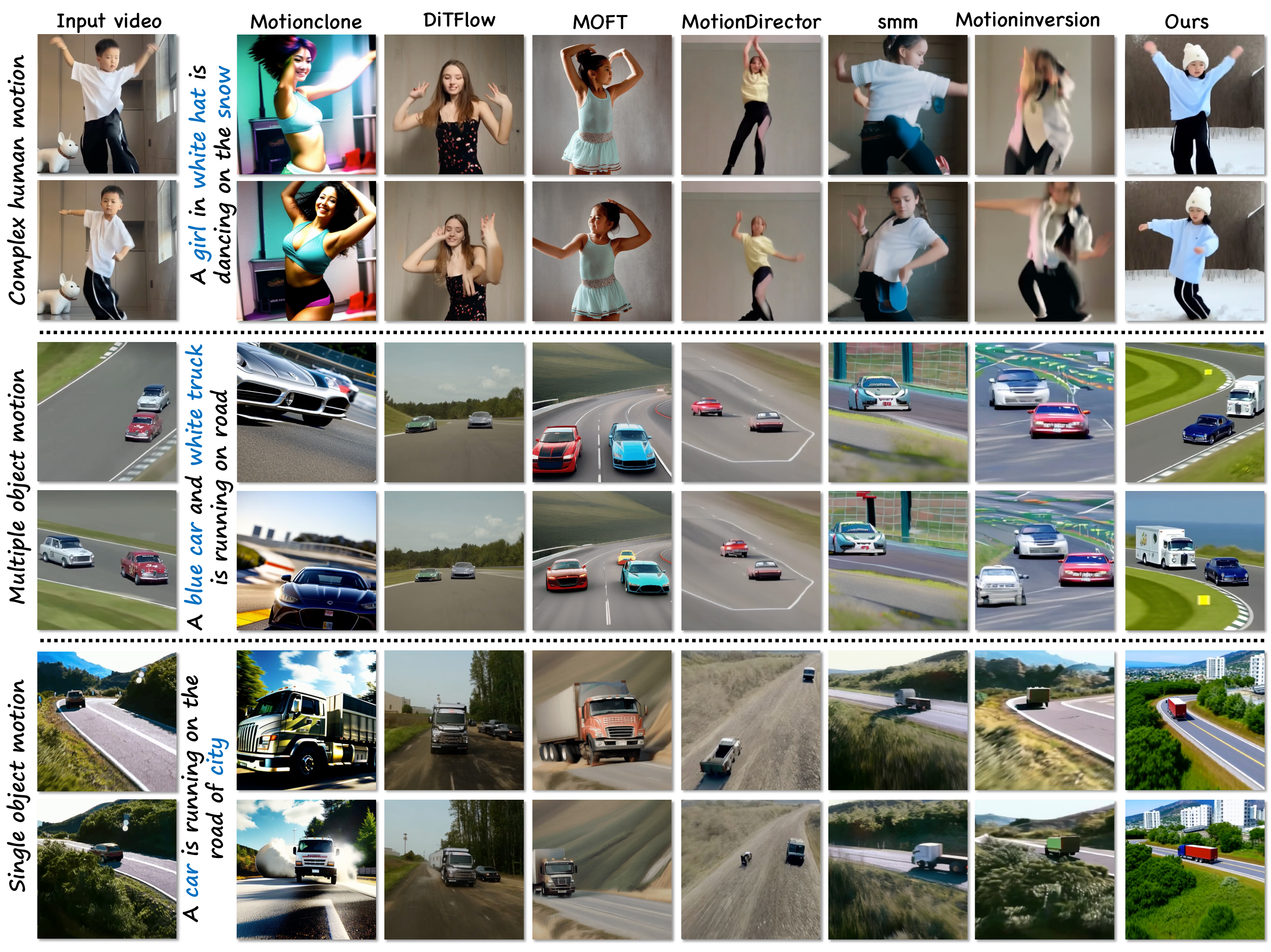}
  \caption{\textbf{Qualitative comparison with baselines.} We perform the visual comparison with various baselines using various kinds of motions. Our method obtains better performance in various motions.}
  % \vspace{-0.6cm}
  \label{fig:comparison}
\end{figure*}

\subsection{MotionBench}

In order to address the lack of a benchmark in video motion transfer, we introduce MotionBench, a comprehensive benchmark to evaluate the ability of current motion transfer approaches. In detail, we collect 200 videos from four aspects, including 1). camera motion, 2). single object motion, 3). multiple object motion, and 4). complex human motion. 
Single object motion sequences focus on diverse motion patterns from a single subject. 
Multiple object motion involves the consistency of spatial relationships between different instances.
Camera motion evaluates viewpoint changes through both simple camera trajectories (zoom, tilt, pan) and complex camera operations.
Here, single/multi-object refers to general objects and animals, while human motion contains more non-rigid deformations, so we treat it separately.
The 30\% videos in our benchmark are generated by text-to-video generation models~\citep{wan2025}, and other videos are obtained from publicly licensed video websites.
We use the GPT4o~\citep{openai2024gpt4o} to get the video captions. In MotionBench, each video is approximately 5 seconds long with 150 frames.
MotionBench provides a standardized evaluation protocol across diverse motion categories, enabling systematic assessment and comparison of motion transfer methods.

\subsection{Comparison with baselines}
\label{sec: comparison}

\begin{table}[t]
\centering
\caption{\small \textbf{Comparison with state-of-the-art video motion transfer methods}. \textcolor{Red}{\textbf{Red}} and \textcolor{Blue}{\textbf{Blue}} denote the best and second best results, respectively.}
\resizebox{\columnwidth}{!}{
\centering
\begin{tabular}{c|cccc|cccc}
\toprule
\multirow{2}{*}{Method}
& \multicolumn{4}{c|}{Quantitative Metrics}
& \multicolumn{4}{c}{User Study} \\
\cmidrule(lr){2-5}\cmidrule(lr){6-9}
& Text Sim.↑ & Motion Fid.↑ & Temp. Cons.↑ & Time (s)↓
& Motion Pres.↓ & App.↓ & Text Align.↓ & Overall↓ \\
\midrule
\multicolumn{9}{c}{\textbf{Training-free methods}} \\
\midrule
MOFT~\citep{xiao2024video}
& 0.286 & 0.792 & 0.922 & 1230
& 6.883 & 6.631 & 5.894 & 6.639 \\
MotionClone~\citep{ling2024motionclone}
 & 0.302 & 0.831 & 0.901 & 1015
& 6.283 & 5.874 & 6.642 & 4.192 \\
SMM~\citep{yatim2024space}
& 0.279 & \textbf{\textcolor{Blue}{0.932}} & 0.918 & 775
& 4.350 & 5.086 & 4.205 & 5.883 \\
DiTFlow~\citep{pondaven2025ditflow}
& \textbf{\textcolor{Blue}{0.375}} & 0.807 &\textbf{\textcolor{Blue}{0.941}} & \textbf{\textcolor{Red}{712}}
& 3.326 & \textbf{\textcolor{Blue}{2.417}} & \textbf{\textcolor{Blue}{2.215}} & 3.284 \\
\midrule
\multicolumn{9}{c}{\textbf{Tuning-based methods}} \\
\midrule
MotionInversion~\citep{jeong2024vmc}
& 0.295 & 0.831 & 0.771 & 2315
& 5.417 & 3.295 & 5.117 & 5.074 \\
MotionDirector~\citep{zhao2023motiondirector}
 & 0.292 & 0.896 & 0.939 & 3008
& \textbf{\textcolor{Blue}{2.217}} & 4.208 & 3.298 & \textbf{\textcolor{Blue}{2.216}} \\

Ours
 & \textbf{\textcolor{Red}{0.380}} & \textbf{\textcolor{Red}{0.971}} & \textbf{\textcolor{Red}{0.976}} & \textbf{\textcolor{Blue}{727}} 
& \textbf{\textcolor{Red}{1.123}} & \textbf{\textcolor{Red}{1.335}} & \textbf{\textcolor{Red}{1.174}} & \textbf{\textcolor{Red}{1.132}} \\
\bottomrule
\end{tabular}%
}
\captionsetup{font=small}
 
% \vspace{-10mm}
\label{tab:comparison}

\end{table}

% To obtain a fairassessment, similar to the approach used in DiTFlow~\citep{pondaven2024videomotion} ,  we adapt   MOFT~\citep{xiao2024video}, MotionInversion~\citep{wang2024motioninversion}, MotionClone~\citep{ling2024motionclone}, SMM~\citep{yatim2024space}, MotionDirector~\citep{zhao2023motiondirector}, DiTFlow~\citep{pondaven2024videomotion} to Wan2.1.

In the following paragraphs, we qualitatively and quantitatively compare our method with previous state-of-the-art methods. We also apply their methods to Wan-2.1~\citep{wan2025} and CogVideo~\citep{yang2024cogvideox} for fair comparison. 

\noindent\textbf{Qualitative comparison.}
We compare our approach with previous video motion transfer methods visually, including state-of-the-art video motion transfer methods: MOFT~\citep{xiao2024video}, MotionInversion~\citep{wang2024motioninversion}, MotionClone~\citep{ling2024motionclone}, SMM~\citep{yatim2024space}, MotionDirector~\citep{zhao2023motiondirector}, DiTFlow~\citep{pondaven2024videomotion}.  
We exclude Motionshop~\citep{yesiltepe2024motionshop} and MotionCrafter~\citep{zhang2023motioncrafter} from our comparisons as no public release exists.
Our experimental results exhibit \ours~better performance and versatility across diverse motion transfer scenarios. 
As illustrated in Fig. \ref{fig:comparison}, in single object motion cases (first column), we find that the previous works fail to follow source motion. 
In contrast, our approaches effectively transform the motion from the source video into the target object, maintaining a consistent motion pattern. 
For multi-object cases, MotionDirector~\citep{zhao2023motiondirector} and SMM~\citep{yatim2024space} have the challenge of handling multi-object interaction motion. 
Our method enables generating videos with aligned movement patterns, preserving the spatial relationships between moving subjects. Additionally, we provide a visual comparison of complex camera motion. The visual results demonstrate the superiority of our methods in camera motion transfer capabilities. 
% See more results in the supplementary materials.

% 

\noindent\textbf{Quantitative comparison.} 
We compare our method with state-of-the-art video motion transfer on our MotionBench, and the results are shown in Tab.~\ref{tab:comparison}. Due to the limited video length of previous works, all evaluations are performed in 32 frames at a resolution of $512 \times 512$. 
Here, we classify the SOTA methods as two classes, training-free or tuning-based, according to whether they use spatial/temporal LoRA to optimize complex motion patterns.
(a) \textbf{Time}: Thanks to sparse motion sampling, \ours~is the fastest tuning-based method. Moreover, our running time is on par with training-free approaches while delivering superior performance.
(b) \textbf{Motion Fidelity}: Following ~\citep{yatim2024space}, motion fidelity is applied to evaluate tracklet similarity between reference and output videos. 
(c) \textbf{Temporal Consistency}: We evaluate the average frame-to-frame coherence using CLIP~\citep{radford2021learning}  feature similarity among consecutive video frames.
(d)\textbf{Text similarity}: We use CLIP to extract target video features and compute the average cosine similarity between the input prompt and all video frames.
(f) \textbf{User study}: Since automatic metrics often fail to reflect real preferences, we invited 20 volunteers to rank methods on MotionBench across four aspects including motion preservation, appearance diversity, text alignment, and overall quality from 1 (best) to 7(worst). The average rank per method (lower is better) is shown in Tab.~\ref{tab:comparison} (1=best, 7=worst). Our method achieves the top result in both automatic metrics and human preference.

\subsection{Ablation study}

\label{sec:ablation}
\begin{wrapfigure}{tr}{0.6\textwidth}  % r：靠右，宽度为 0.5\textwidth
  \vspace{-2em}
  \centering

  \renewcommand{\arraystretch}{1.2}
  \definecolor{Red}{RGB}{192,0,0}
  \definecolor{Blue}{RGB}{12,114,186}
  % —— 下半：表格 —— %
  \begin{minipage}{\linewidth}
    \centering
    \renewcommand{\arraystretch}{1.1} % 适当调小行高
    \captionof{table}{\footnotesize \textbf{Quantitative ablation}. \textcolor{Red}{\textbf{Red}} and \textcolor{Blue}{\textbf{Blue}}  denote best, 2nd. \textcolor{black}{ Baseline means we disable all three proposed components simultaneously.} }
    \vspace{-1mm}
    {\scriptsize                      % ← 用 scriptsize，甚至可以试试 \tiny
      \begin{tabular}{@{}l|cccc@{}}
        \toprule
        Method & Text Sim.$\uparrow$ & Motion Fid.$\uparrow$ & Temp. Cons.$\uparrow$ & Time(s)$\downarrow$ \\
        \midrule
        \textcolor{black}{Baseline}     & \textcolor{black}{0.362} & \textcolor{black}{0.658} & \textcolor{black}{0.824} & \textcolor{black}{2493}  \\
        w/o STD LoRa        & 0.364 & 0.546 & 0.845 & 971  \\
        w/o Adaptive RoPE   & 0.371 & 0.655 & 0.817 & \textcolor{Blue}{\textbf{792}}  \\
        w/o Sparse Sampling & \textcolor{Blue}{\textbf{0.369}} & \textcolor{Red}{\textbf{0.975}} & \textcolor{Blue}{\textbf{0.967}} & 2068 \\
        \midrule
        Ours & \textcolor{Red}{\textbf{0.380}} & \textcolor{Blue}{\textbf{0.971}} & \textcolor{Red}{\textbf{0.976}} & \textcolor{Red}{\textbf{727}} \\
        \bottomrule
      \end{tabular}
    }

    \label{tab:ablation}

  \end{minipage}

  % —— 上半：图片 —— %
  \begin{minipage}{\linewidth}
    \centering
    \includegraphics[width=\linewidth]{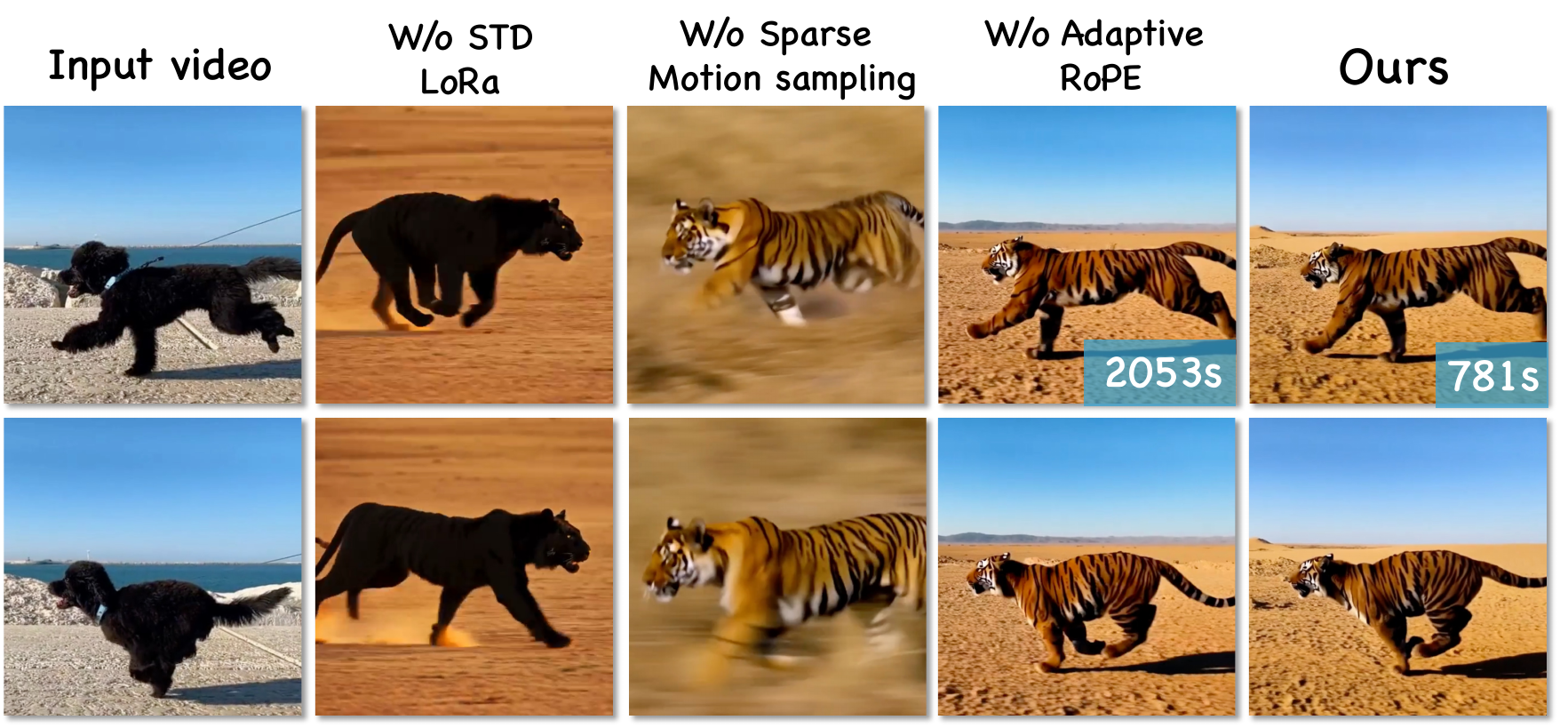}
    \captionof{figure}{\textbf{Ablation study about proposed modules}. We remove the proposed modules to evaluate their effectiveness. “STD” means spatial–temporal decoupled LoRA.}
    \label{fig:abla_merge}
  \end{minipage}
  \vspace{1ex}

\vspace{-6mm}

\end{wrapfigure}

In this section, we conduct a systematic ablation study to isolate and quantify the contribution of each key component in our framework. The qualitative and quantitative ablation study results are shown in Fig.~\ref{fig:abla_merge} and Tab.~\ref{tab:ablation}, respectively. 
% More ablation studies can be found in the Appendix.
%Specifically, we first verify the effectiveness of the spatial–temporal decoupled LoRA module; next, we evaluate the impact of our sparse motion sampling strategy; and finally, we analyze the benefits brought by the adaptive RoPE mechanism.

\noindent\textbf{Effectiveness of spatial–temporal decoupled LoRA.}  
As shown in the second row of Fig.~\ref{fig:abla_merge} and the “w/o STD LoRA” ablation in Tab.~\ref{tab:ablation}, the naive baseline jointly tunes without separating spatial and temporal attention heads, failing to decouple the dog’s appearance and causing the edited tiger to look unnaturally black. In contrast, our decoupled LoRA preserves motion while effectively modifying appearance, as evidenced by the improved text similarity in Tab.~\ref{tab:ablation}.

\noindent\textbf{Effectiveness of adaptive RoPE.}
Thanks to our adaptive RoPE design, the model can precisely infer each sampled frame’s original index under sparse motion sampling, ensuring the edited motion remains aligned with the source. Without adaptive RoPE, the tiger’s motion becomes disordered and fails to match the original video dynamics. In Tab.~\ref{tab:ablation}, an improvement of about 48.3\% over motion fidelity, quantitatively confirms the benefit of our adaptive RoPE.

\noindent\textbf{Effectiveness of sparse motion sampling.}
By employing sparse motion sampling in the temporal tuning phase, we reduce the tuning time to 727s. Note that in the “w/o sparse sampling” setting, we still apply adaptive RoPE but tune on all video frames, resulting in identical motion fidelity {(0.975 \textit{vs.} 0.971)} while incurring the higher time cost.

% \begin{minipage}{0.6\textwidth}
%     \centering
%     \includegraphics[width=\linewidth]{images/ablation.pdf} % 插入图片
%     % \vspace{-5mm}
%     \captionof{figure}{\textbf{Ablation study about proposed modules}. We remove the proposed modules to evaluate their effectiveness, representatively. \enquote{STD} means spatial-temporal decoupled LoRA. } % 为图片添加标题
%     \label{fig:abla_merge}
% \end{minipage}%

% \begin{minipage}{0.5\textwidth}
%     \centering
%     \includegraphics[width=\linewidth]{images/ablation_STD.pdf} % 插入图片
%     % \vspace{-5mm}
%     \captionof{figure}{Ablation study about random sampling.} % 为图片添加标题
%     \label{fig:abla_merge}

% \end{minipage}%

% \section{Limitation}

\section{Conclusion}

% In this paper, we propose COVE, which is the first to explore how to employ inherent diffusion feature correspondence in video editing to enhance editing quality and temporal consistency. 
% Through the proposed efficient sliding-window-based strategy, the one-to-many correspondence relationship among tokens across frames is obtained. During the inversion and denoising process, self-attention is performed within the corresponding tokens to enhance temporal consistency. Additionally, we also apply token merging in the temporal dimension to improve the efficiency of the editing process. Both quantitative and qualitative experimental results demonstrate the effectiveness of our method, which outperforms a wide range of previous methods, achieving state-of-the-art editing quality. 

In this paper, we propose \ours,  a three-stage video motion transfer framework that tunes the video diffusion transformer to synthesize video clips with complex motion. In detail, we analyze the motion inefficiency and tuning inefficiency in DiT-based video motion transfer. Through the proposed efficient spatial-temporal decoupled LoRA, we achieve better motion consistency. To address the tuning inefficiency, we introduce adaptive RoPE and sparse motion sampling to accelerate training. Extensive experimental results demonstrate the effectiveness of our method, which outperforms a wide range of previous methods, achieving state-of-the-art video motion transfer quality.

\bibliography{iclr2026_conference}

@article{wang2026care,
  title={CARE-Edit: Condition-Aware Routing of Experts for Contextual Image Editing},
  author={Wang, Yucheng and Wang, Zedong and Wu, Yuetong and Ma, Yue and Xu, Dan},
  journal={arXiv preprint arXiv:2603.08589},
  year={2026}
}

@article{ma2026fastvmt,
  title={FastVMT: Eliminating Redundancy in Video Motion Transfer},
  author={Ma, Yue and Wang, Zhikai and Ren, Tianhao and Zheng, Mingzhe and Liu, Hongyu and Guo, Jiayi and Fong, Mark and Xue, Yuxuan and Zhao, Zixiang and Schindler, Konrad and others},
  journal={arXiv preprint arXiv:2602.05551},
  year={2026}
}

@article{yang2025unified,
  title={Unified Video Editing with Temporal Reasoner},
  author={Yang, Xiangpeng and Xie, Ji and Yang, Yiyuan and Huang, Yan and Xu, Min and Wu, Qiang},
  journal={arXiv preprint arXiv:2512.07469},
  year={2025}
}

@article{shen2025follow,
  title={Follow-Your-Preference: Towards Preference-Aligned Image Inpainting},
  author={Shen, Yutao and Yuan, Junkun and Aonishi, Toru and Nakayama, Hideki and Ma, Yue},
  journal={arXiv preprint arXiv:2509.23082},
  year={2025}
}

@article{ma2026follow,
  title={Follow-your-emoji-faster: Towards efficient, fine-controllable, and expressive freestyle portrait animation},
  author={Ma, Yue and Yan, Zexuan and Liu, Hongyu and Wang, Hongfa and Pan, Heng and He, Yingqing and Yuan, Junkun and Zeng, Ailing and Cai, Chengfei and Shum, Heung-Yeung and others},
  journal={International Journal of Computer Vision},
  volume={134},
  number={3},
  pages={130},
  year={2026},
  publisher={Springer}
}

@article{long2025follow,
  title={Follow-your-shape: Shape-aware image editing via trajectory-guided region control},
  author={Long, Zeqian and Zheng, Mingzhe and Feng, Kunyu and Zhang, Xinhua and Liu, Hongyu and Yang, Harry and Zhang, Linfeng and Chen, Qifeng and Ma, Yue},
  journal={arXiv preprint arXiv:2508.08134},
  year={2025}
}

@article{ma2025controllable,
  title={Controllable video generation: A survey},
  author={Ma, Yue and Feng, Kunyu and Hu, Zhongyuan and Wang, Xinyu and Wang, Yucheng and Zheng, Mingzhe and Wang, Bingyuan and Wang, Qinghe and He, Xuanhua and Wang, Hongfa and others},
  journal={arXiv preprint arXiv:2507.16869},
  year={2025}
}

@article{lv2025bm,
  title={BM-Edit: Background retention and motion consistency for zero-shot video editing},
  author={Lv, Xiang and Shao, Mingwen and Wan, Yecong and Ma, Yue and Cheng, Yuanshuo and Meng, Lingzhuang},
  journal={Knowledge-Based Systems},
  volume={324},
  pages={113784},
  year={2025},
  publisher={Elsevier}
}

@article{yang2025evctrl,
  title={Evctrl: Efficient control adapter for visual generation},
  author={Yang, Zixiang and Ma, Yue and Zhang, Yinhan and Mo, Shanhui and Liu, Dongrui and Zhang, Linfeng},
  journal={arXiv preprint arXiv:2508.10963},
  year={2025}
}

@article{chen2025contextflow,
  title={ContextFlow: Training-Free Video Object Editing via Adaptive Context Enrichment},
  author={Chen, Yiyang and He, Xuanhua and Ma, Xiujun and Ma, Yue},
  journal={arXiv preprint arXiv:2509.17818},
  year={2025}
}

@inproceedings{feng2025dit4edit,
  title={Dit4edit: Diffusion transformer for image editing},
  author={Feng, Kunyu and Ma, Yue and Wang, Bingyuan and Qi, Chenyang and Chen, Haozhe and Chen, Qifeng and Wang, Zeyu},
  booktitle={Proceedings of the AAAI Conference on Artificial Intelligence},
  volume={39},
  number={3},
  pages={2969--2977},
  year={2025}
}

@article{liu2025longvideoagent,
  title={LongVideoAgent: Multi-Agent Reasoning with Long Videos},
  author={Liu, Runtao and Liu, Ziyi and Tang, Jiaqi and Ma, Yue and Pi, Renjie and Zhang, Jipeng and Chen, Qifeng},
  journal={arXiv preprint arXiv:2512.20618},
  year={2025}
}

@inproceedings{zhang2025magiccolor,
  title={Magiccolor: Multi-instance sketch colorization},
  author={Zhang, Yinhan and Ma, Yue and Wang, Bingyuan and Chen, Qifeng and Wang, Zeyu},
  booktitle={Proceedings of the IEEE/CVF International Conference on Computer Vision},
  pages={15205--15217},
  year={2025}
}

@article{zhu2024instantswap,
  title={Instantswap: Fast customized concept swapping across sharp shape differences},
  author={Zhu, Chenyang and Li, Kai and Ma, Yue and Tang, Longxiang and Fang, Chengyu and Chen, Chubin and Chen, Qifeng and Li, Xiu},
  journal={arXiv preprint arXiv:2412.01197},
  year={2024}
}

@article{hu2024motionmaster,
  title={Motionmaster: Training-free camera motion transfer for video generation},
  author={Hu, Teng and Zhang, Jiangning and Yi, Ran and Wang, Yating and Huang, Hongrui and Weng, Jieyu and Wang, Yabiao and Ma, Lizhuang},
  journal={arXiv preprint arXiv:2404.15789},
  year={2024}
}

@article{abdal2025dynamic,
  title={Dynamic Concepts Personalization from Single Videos},
  author={Abdal, Rameen and Patashnik, Or and Skorokhodov, Ivan and Menapace, Willi and Siarohin, Aliaksandr and Tulyakov, Sergey and Cohen-Or, Daniel and Aberman, Kfir},
  journal={arXiv preprint arXiv:2502.14844},
  year={2025}
}

@article{zhang2025framepainter,
  title={FramePainter: Endowing Interactive Image Editing with Video Diffusion Priors},
  author={Zhang, Yabo and Zhou, Xinpeng and Zeng, Yihan and Xu, Hang and Li, Hui and Zuo, Wangmeng},
  journal={arXiv preprint arXiv:2501.08225},
  year={2025}
}

@article{xing2024make,
  title={Make-your-video: Customized video generation using textual and structural guidance},
  author={Xing, Jinbo and Xia, Menghan and Liu, Yuxin and Zhang, Yuechen and Zhang, Yong and He, Yingqing and Liu, Hanyuan and Chen, Haoxin and Cun, Xiaodong and Wang, Xintao and others},
  journal={IEEE Transactions on Visualization and Computer Graphics},
  year={2024},
  publisher={IEEE}
}

@article{vaswani2017attention,
  title={Attention is all you need},
  author={Vaswani, Ashish and Shazeer, Noam and Parmar, Niki and Uszkoreit, Jakob and Jones, Llion and Gomez, Aidan N and Kaiser, {\L}ukasz and Polosukhin, Illia},
  journal={Advances in neural information processing systems},
  volume={30},
  year={2017}
}

@inproceedings{xing2024dynamicrafter,
  title={Dynamicrafter: Animating open-domain images with video diffusion priors},
  author={Xing, Jinbo and Xia, Menghan and Zhang, Yong and Chen, Haoxin and Yu, Wangbo and Liu, Hanyuan and Liu, Gongye and Wang, Xintao and Shan, Ying and Wong, Tien-Tsin},
  booktitle={European Conference on Computer Vision},
  pages={399--417},
  year={2024},
  organization={Springer}
}

@article{chen2025s,
  title={S-Guidance: Stochastic Self Guidance for Training-Free Enhancement of Diffusion Models},
  author={Chen, Chubin and Zhu, Jiashu and Feng, Xiaokun and Huang, Nisha and Wu, Meiqi and Mao, Fangyuan and Wu, Jiahong and Chu, Xiangxiang and Li, Xiu},
  journal={arXiv preprint arXiv:2508.12880},
  year={2025}
}

@article{chen2025taming,
  title={Taming Preference Mode Collapse via Directional Decoupling Alignment in Diffusion Reinforcement Learning},
  author={Chen, Chubin and Hu, Sujie and Zhu, Jiashu and Wu, Meiqi and Chen, Jintao and Li, Yanxun and Huang, Nisha and Fang, Chengyu and Wu, Jiahong and Chu, Xiangxiang and others},
  journal={arXiv preprint arXiv:2512.24146},
  year={2025}
}

@article{xu2024easyanimate,
  title={Easyanimate: A high-performance long video generation method based on transformer architecture},
  author={Xu, Jiaqi and Zou, Xinyi and Huang, Kunzhe and Chen, Yunkuo and Liu, Bo and Cheng, MengLi and Shi, Xing and Huang, Jun},
  journal={arXiv preprint arXiv:2405.18991},
  year={2024}
}

@article{kong2024hunyuanvideo,
  title={Hunyuanvideo: A systematic framework for large video generative models},
  author={Kong, Weijie and Tian, Qi and Zhang, Zijian and Min, Rox and Dai, Zuozhuo and Zhou, Jin and Xiong, Jiangfeng and Li, Xin and Wu, Bo and Zhang, Jianwei and others},
  journal={arXiv preprint arXiv:2412.03603},
  year={2024}
}

@inproceedings{rombach2022high,
  title={High-resolution image synthesis with latent diffusion models},
  author={Rombach, Robin and Blattmann, Andreas and Lorenz, Dominik and Esser, Patrick and Ommer, Bj{\"o}rn},
  booktitle={Proceedings of the IEEE/CVF conference on computer vision and pattern recognition},
  pages={10684--10695},
  year={2022}
}

@inproceedings{guoanimatediff,
  title={AnimateDiff: Animate Your Personalized Text-to-Image Diffusion Models without Specific Tuning},
  author={Guo, Yuwei and Yang, Ceyuan and Rao, Anyi and Liang, Zhengyang and Wang, Yaohui and Qiao, Yu and Agrawala, Maneesh and Lin, Dahua and Dai, Bo},
  booktitle={The Twelfth International Conference on Learning Representations},
  year={2024}
}

@article{wang2023modelscope,
  title={Modelscope text-to-video technical report},
  author={Wang, Jiuniu and Yuan, Hangjie and Chen, Dayou and Zhang, Yingya and Wang, Xiang and Zhang, Shiwei},
  journal={arXiv preprint arXiv:2308.06571},
  year={2023}
}

@article{wu2022tune,
  title={Tune-a-video: One-shot tuning of image diffusion models for text-to-video generation},
  author={Wu, Jay Zhangjie and Ge, Yixiao and Wang, Xintao and Lei, Weixian and Gu, Yuchao and Hsu, Wynne and Shan, Ying and Qie, Xiaohu and Shou, Mike Zheng},
  journal={arXiv preprint arXiv:2212.11565},
  year={2022}
}

@article{qi2023fatezero,
  title={Fatezero: Fusing attentions for zero-shot text-based video editing},
  author={Qi, Chenyang and Cun, Xiaodong and Zhang, Yong and Lei, Chenyang and Wang, Xintao and Shan, Ying and Chen, Qifeng},
  journal={arXiv preprint arXiv:2303.09535},
  year={2023}
}

@article{zhao2023controlvideo,
  title={ControlVideo: Adding Conditional Control for One Shot Text-to-Video Editing},
  author={Zhao, Min and Wang, Rongzhen and Bao, Fan and Li, Chongxuan and Zhu, Jun},
  journal={arXiv preprint arXiv:2305.17098},
  year={2023}
}

@article{geyer2023tokenflow,
  title={Tokenflow: Consistent diffusion features for consistent video editing},
  author={Geyer, Michal and Bar-Tal, Omer and Bagon, Shai and Dekel, Tali},
  journal={arXiv preprint arXiv:2307.10373},
  year={2023}
}

@article{zhao2023motiondirector,
  title={MotionDirector: Motion Customization of Text-to-Video Diffusion Models},
  author={Zhao, Rui and Gu, Yuchao and Wu, Jay Zhangjie and Zhang, David Junhao and Liu, Jiawei and Wu, Weijia and Keppo, Jussi and Shou, Mike Zheng},
  journal={arXiv preprint arXiv:2310.08465},
  year={2023}
}

@article{guo2023animatediff,
  title={Animatediff: Animate your personalized text-to-image diffusion models without specific tuning},
  author={Guo, Yuwei and Yang, Ceyuan and Rao, Anyi and Wang, Yaohui and Qiao, Yu and Lin, Dahua and Dai, Bo},
  journal={arXiv preprint arXiv:2307.04725},
  year={2023}
}

@inproceedings{radford2021learning,
  title={Learning transferable visual models from natural language supervision},
  author={Radford, Alec and Kim, Jong Wook and Hallacy, Chris and Ramesh, Aditya and Goh, Gabriel and Agarwal, Sandhini and Sastry, Girish and Askell, Amanda and Mishkin, Pamela and Clark, Jack and others},
  booktitle={International conference on machine learning},
  pages={8748--8763},
  year={2021},
  organization={PMLR}
}

@inproceedings{yatim2024space,
  title={Space-time diffusion features for zero-shot text-driven motion transfer},
  author={Yatim, Danah and Fridman, Rafail and Bar-Tal, Omer and Kasten, Yoni and Dekel, Tali},
  booktitle={Proceedings of the IEEE/CVF Conference on Computer Vision and Pattern Recognition},
  pages={8466--8476},
  year={2024}
}

@article{jeong2024dreammotion,
  title={DreamMotion: Space-Time Self-Similarity Score Distillation for Zero-Shot Video Editing},
  author={Jeong, Hyeonho and Chang, Jinho and Park, Geon Yeong and Ye, Jong Chul},
  journal={arXiv preprint arXiv:2403.12002},
  year={2024}
}

@article{ma2024followyouremoji,
title={Follow-Your-Emoji: Fine-Controllable and Expressive Freestyle Portrait Animation},
author={Ma, Yue and Liu, Hongyu and Wang, Hongfa and Pan, Heng and He, Yingqing and Yuan, Junkun and Zeng, Ailing and Cai, Chengfei and Shum, Heung-Yeung and Liu, Wei and others},
journal={arXiv preprint arXiv:2406.01900},
year={2024}
}

@inproceedings{ma2024follow,
  title={Follow your pose: Pose-guided text-to-video generation using pose-free videos},
  author={Ma, Yue and He, Yingqing and Cun, Xiaodong and Wang, Xintao and Chen, Siran and Li, Xiu and Chen, Qifeng},
  booktitle={Proceedings of the AAAI Conference on Artificial Intelligence},
  pages={4117--4125},
  year={2024}
}

@article{ma2024followyourclick,   
title={Follow-Your-Click: Open-domain Regional Image Animation via Short Prompts},   
author={Ma, Yue and He, Yingqing and Wang, Hongfa and Wang, Andong and Qi, Chenyang and Cai, Chengfei and Li, Xiu and Li, Zhifeng and Shum, Heung-Yeung and Liu, Wei and others},  
journal={arXiv preprint arXiv:2403.08268},   
year={2024} 
}

@article{chen2024follow,
  title={Follow-your-canvas: Higher-resolution video outpainting with extensive content generation},
  author={Chen, Qihua and Ma, Yue and Wang, Hongfa and Yuan, Junkun and Zhao, Wenzhe and Tian, Qi and Wang, Hongmei and Min, Shaobo and Chen, Qifeng and Liu, Wei},
  journal={arXiv preprint arXiv:2409.01055},
  year={2024}
}

@article{wang2024cove,
  title={COVE: Unleashing the Diffusion Feature Correspondence for Consistent Video Editing},
  author={Wang, Jiangshan and Ma, Yue and Guo, Jiayi and Xiao, Yicheng and Huang, Gao and Li, Xiu},
  journal={arXiv preprint arXiv:2406.08850},
  year={2024}
}

@article{ma2023magicstick,
  title={Magicstick: Controllable video editing via control handle transformations},
  author={Ma, Yue and Cun, Xiaodong and He, Yingqing and Qi, Chenyang and Wang, Xintao and Shan, Ying and Li, Xiu and Chen, Qifeng},
  journal={arXiv preprint arXiv:2312.03047},
  year={2023}
}

@article{yang2024cogvideox,
  title={Cogvideox: Text-to-video diffusion models with an expert transformer},
  author={Yang, Zhuoyi and Teng, Jiayan and Zheng, Wendi and Ding, Ming and Huang, Shiyu and Xu, Jiazheng and Yang, Yuanming and Hong, Wenyi and Zhang, Xiaohan and Feng, Guanyu and others},
  journal={arXiv preprint arXiv:2408.06072},
  year={2024}
}

@inproceedings{pondaven2025ditflow,
      title={Video Motion Transfer with Diffusion Transformers}, 
      author={Alexander Pondaven and Aliaksandr Siarohin and Sergey Tulyakov and Philip Torr and Fabio Pizzati},
      booktitle={CVPR},
      year={2025}
}

@article{xi2025sparse,
  title={Sparse VideoGen: Accelerating Video Diffusion Transformers with Spatial-Temporal Sparsity},
  author={Xi, Haocheng and Yang, Shuo and Zhao, Yilong and Xu, Chenfeng and Li, Muyang and Li, Xiuyu and Lin, Yujun and Cai, Han and Zhang, Jintao and Li, Dacheng and others},
  journal={arXiv preprint arXiv:2502.01776},
  year={2025}
}

@article{yang2025videograin,
  title={Videograin: Modulating space-time attention for multi-grained video editing},
  author={Yang, Xiangpeng and Zhu, Linchao and Fan, Hehe and Yang, Yi},
  journal={arXiv preprint arXiv:2502.17258},
  year={2025}
}

@article{loshchilov2017decoupled,
  title={Decoupled Weight Decay Regularization},
  author={Loshchilov, Ilya and Hutter, Frank},
  journal={arXiv preprint arXiv:1711.05101},
  year={2017}
}

@techreport{openai2024gpt4o,
  title        = {{GPT-4o} Technical Report},
  author       = {{OpenAI}},
  institution  = {OpenAI},
  year         = {2024},
  url          = {https://chatgpt.com/},
  note         = {Accessed: 2025-05-12}
}

@article{xiao2024video,
  title={Video diffusion models are training-free motion interpreter and controller},
  author={Xiao, Zeqi and Zhou, Yifan and Yang, Shuai and Pan, Xingang},
  journal={Advances in Neural Information Processing Systems},
  volume={37},
  pages={76115--76138},
  year={2024}
}

@inproceedings{wang2024motioninversion,
  title={Motion inversion for video customization},
  author={Wang, Luozhou and Mai, Ziyang and Shen, Guibao and Liang, Yixun and Tao, Xin and Wan, Pengfei and Zhang, Di and Li, Yijun and Chen, Ying-Cong},
  booktitle={Proceedings of the Special Interest Group on Computer Graphics and Interactive Techniques Conference Conference Papers},
  pages={1--12},
  year={2025}
}

@article{ling2024motionclone,
  title={Motionclone: Training-free motion cloning for controllable video generation},
  author={Ling, Pengyang and Bu, Jiazi and Zhang, Pan and Dong, Xiaoyi and Zang, Yuhang and Wu, Tong and Chen, Huaian and Wang, Jiaqi and Jin, Yi},
  journal={arXiv preprint arXiv:2406.05338},
  year={2024}
}

@inproceedings{pondaven2024videomotion,
  title={Video motion transfer with diffusion transformers},
  author={Pondaven, Alexander and Siarohin, Aliaksandr and Tulyakov, Sergey and Torr, Philip and Pizzati, Fabio},
  booktitle={Proceedings of the Computer Vision and Pattern Recognition Conference},
  pages={22911--22921},
  year={2025}
}

@article{yesiltepe2024motionshop,
  title   = {MotionShop: Zero-Shot Motion Transfer in Video Diffusion Models with Mixture of Score Guidance},
  author  = {Yesiltepe, Hidir and Meral, Tuna Han Salih and Dunlop, Connor and Yanardag, Pinar},
  journal = {arXiv preprint arXiv:2412.05355},
  year    = {2024}
}

@article{zhang2023motioncrafter,
  title   = {MotionCrafter: One-Shot Motion Customization of Diffusion Models},
  author  = {Zhang, Yuxin and Tang, Fan and Huang, Nisha and Huang, Haibin and Ma, Chongyang and Dong, Weiming and Xu, Changsheng},
  journal = {arXiv preprint arXiv:2312.05288},
  year    = {2023}
}

@InProceedings{jeong2024vmc,
  author    = {Jeong, Hyeonho and Park, Geon Yeong and Ye, Jong Chul},
  title     = {VMC: Video Motion Customization using Temporal Attention Adaption for Text-to-Video Diffusion Models},
  booktitle = {Proceedings of the IEEE/CVF Conference on Computer Vision and Pattern Recognition (CVPR)},
  month     = {June},
  year      = {2024},
  pages     = {9212--9221},
}

@article{Liu2024SoraAR,
  title={Sora: A review on background, technology, limitations, and opportunities of large vision models},
  author={Liu, Yixin and Zhang, Kai and Li, Yuan and Yan, Zhiling and Gao, Chujie and Chen, Ruoxi and Yuan, Zhengqing and Huang, Yue and Sun, Hanchi and Gao, Jianfeng and others},
  journal={arXiv preprint arXiv:2402.17177},
  year={2024}
}

@article{Yang2024CogVideoXTD,
  title={Cogvideox: Text-to-video diffusion models with an expert transformer},
  author={Yang, Zhuoyi and Teng, Jiayan and Zheng, Wendi and Ding, Ming and Huang, Shiyu and Xu, Jiazheng and Yang, Yuanming and Hong, Wenyi and Zhang, Xiaohan and Feng, Guanyu and others},
  journal={arXiv preprint arXiv:2408.06072},
  year={2024}
}

@String(CVPR= {IEEE Conf. Comput. Vis. Pattern Recog.})

@String(ICCV= {Int. Conf. Comput. Vis.})

@String(AAAI = {AAAI})

@String(CVPR  = {CVPR})

@String(ICCV  = {ICCV})

@article{xue2024follow,
  title={Follow-Your-Pose v2: Multiple-Condition Guided Character Image Animation for Stable Pose Control},
  author={Xue, Jingyun and Wang, Hongfa and Tian, Qi and Ma, Yue and Wang, Andong and Zhao, Zhiyuan and Min, Shaobo and Zhao, Wenzhe and Zhang, Kaihao and Shum, Heung-Yeung and others},
  journal={arXiv preprint arXiv:2406.03035},
  year={2024}
}

@inproceedings{chai2023stablevideo,
  title={Stablevideo: Text-driven consistency-aware diffusion video editing},
  author={Chai, Wenhao and Guo, Xun and Wang, Gaoang and Lu, Yan},
  booktitle=ICCV,
  year={2023}
}

@inproceedings{liu2025avatarartist,
author    = {Liu, Hongyu and Wang, Xuan and Wan, Ziyu and Ma, Yue and Chen, Jingye and Fan, Yanbo and Shen, Yujun and Song, Yibing and Chen, Qifeng},
title     = {AvatarArtist: Open-Domain 4D Avatarization},
booktitle = {CVPR},
year      = {2025}}

@inproceedings{zhu2025multibooth,
  title={Multibooth: Towards generating all your concepts in an image from text},
  author={Zhu, Chenyang and Li, Kai and Ma, Yue and He, Chunming and Li, Xiu},
  booktitle={Proceedings of the AAAI Conference on Artificial Intelligence},
  volume={39},
  number={10},
  pages={10923--10931},
  year={2025}
}

@inproceedings{chen2024m,
  title={M-bev: Masked bev perception for robust autonomous driving},
  author={Chen, Siran and Ma, Yue and Qiao, Yu and Wang, Yali},
  booktitle={Proceedings of the AAAI Conference on Artificial Intelligence},
  volume={38},
  number={2},
  pages={1183--1191},
  year={2024}
}

@article{chen2023attentive,
  title={Attentive snippet prompting for video retrieval},
  author={Chen, Siran and Xu, Qinglin and Ma, Yue and Qiao, Yu and Wang, Yali},
  journal={IEEE Transactions on Multimedia},
  volume={26},
  pages={4348--4359},
  year={2023},
  publisher={IEEE}
}

@article{wan2025,
  title={Wan: Open and advanced large-scale video generative models},
  author={Wan, Team and Wang, Ang and Ai, Baole and Wen, Bin and Mao, Chaojie and Xie, Chen-Wei and Chen, Di and Yu, Feiwu and Zhao, Haiming and Yang, Jianxiao and others},
  journal={arXiv preprint arXiv:2503.20314},
  year={2025}
}

@article{wang2025multishotmaster,
  title={MultiShotMaster: A Controllable Multi-Shot Video Generation Framework},
  author={Wang, Qinghe and Shi, Xiaoyu and Li, Baolu and Bian, Weikang and Liu, Quande and Lu, Huchuan and Wang, Xintao and Wan, Pengfei and Gai, Kun and Jia, Xu},
  journal={arXiv preprint arXiv:2512.03041},
  year={2025}
}

@inproceedings{wang2025cinemaster,
  title={Cinemaster: A 3d-aware and controllable framework for cinematic text-to-video generation},
  author={Wang, Qinghe and Luo, Yawen and Shi, Xiaoyu and Jia, Xu and Lu, Huchuan and Xue, Tianfan and Wang, Xintao and Wan, Pengfei and Zhang, Di and Gai, Kun},
  booktitle={Proceedings of the Special Interest Group on Computer Graphics and Interactive Techniques Conference Conference Papers},
  pages={1--10},
  year={2025}
}

@article{wang2025characterfactory,
  title={Characterfactory: Sampling consistent characters with gans for diffusion models},
  author={Wang, Qinghe and Li, Baolu and Li, Xiaomin and Cao, Bing and Ma, Liqian and Lu, Huchuan and Jia, Xu},
  journal={IEEE Transactions on Image Processing},
  year={2025},
  publisher={IEEE}
}
\bibliographystyle{iclr2026_conference}

\end{document}